\documentclass{article}

\usepackage{arxiv}

\usepackage[utf8]{inputenc} 
\usepackage[T1]{fontenc}    
\usepackage{hyperref}       
\usepackage{url}            
\usepackage{booktabs}       
\usepackage{amsfonts}       
\usepackage{nicefrac}       
\usepackage{microtype}      
\usepackage{lipsum}
\usepackage{graphicx}
\graphicspath{ {./images/} }

\usepackage{graphicx}
\usepackage{amsmath}
\usepackage{amsfonts}
\usepackage{amssymb}
\usepackage{hyperref}
\usepackage{fancyhdr}
\usepackage{subcaption}
\usepackage{geometry}
\usepackage[english]{babel}
\usepackage{csquotes}
\usepackage{xcolor} 
\usepackage{listings}
\usepackage{soul}
\usepackage{float}
\usepackage[linesnumbered,ruled,vlined]{algorithm2e}
\usepackage{multirow}
\usepackage{verbatim}

\title{Recurrent Deep Kernel Learning\\ of Dynamical Systems}

\author{
 Nicolò Botteghi \\
  University of Twente\\
  Enschede, Netherlands \\
  \texttt{n.botteghi@utwente.nl} \\
   \And
 Paolo Motta \\
  Politecnico di Milano\\
  Milano, Italy\\
  \texttt{paolo9.motta@mail.polimi.it} \\
  \And
 Andrea Manzoni \\
  Politecnico di Milano\\
  Milano, Italy\\
  \texttt{andrea1.manzoni@polimi.it} \\
   \AND
 Paolo Zunino\\
  Politecnico di Milano\\
  Milano, Italy\\
  \texttt{paolo.zunino@polimi.it} \\
   \And
  Mengwu Guo \\
  Lund University \\
  Lund, Sweden \\
  \texttt{mengwu.guo@math.lu.se} \\
}

\begin{document}
\maketitle
\begin{abstract}
Digital twins require computationally-efficient reduced-order models (ROMs) that can accurately describe complex dynamics of physical assets. However, constructing ROMs from noisy high-dimensional data is challenging. In this work, we propose a data-driven, non-intrusive method that utilizes stochastic variational deep kernel learning (SVDKL) to discover low-dimensional latent spaces from data and a recurrent version of SVDKL for representing and predicting the evolution of latent dynamics. The proposed method is demonstrated with two challenging examples -- a double pendulum and a reaction-diffusion system. Results show that our framework is capable of (i) denoising and reconstructing measurements, (ii) learning compact representations of system states, (iii) predicting system evolution in low-dimensional latent spaces, and (iv) quantifying modeling uncertainties. 
\end{abstract}

\section{Introduction}

A digital twin \cite{grieves2014digital, tao2018digital, national2023foundational, willcox2024role} is a virtual replica that mimics the structure, context, and behavior of a physical system. The digital twin is synchronized to its physical twin with data in real time to predict dynamical behaviors and inform critical decisions. Constructing a digital twin typically requires accurate modeling of complex physical phenomena that are often described by nonlinear, time-dependent,  parametric partial differential equations (PDEs). Full-order models (FOMs) generated by detailed solvers are often computationally demanding and thus not suitable for the multi-query, real-time contexts in digital twinning, such as uncertainty quantification \cite{sudret2000stochastic, bui2008parametric, galbally2010non}, optimal control \cite{ravindran2000reduced, negri2013reduced, troltzsch2024optimal}, shape optimization \cite{manzoni2012shape}, parameter estimation \cite{cui2015data, frangos2010surrogate}, and model calibration \cite{cao2007reduced, maday2015parameterized}. Therefore, constructing computationally efficient reduced-order models (ROMs) with controlled accuracy is crucial for dealing with real-world systems \cite{quarteroni2015reduced}.

Generally speaking, any surrogate model that reduces the computational cost of a FOM can be considered a ROM. ROMs aim to intelligently represent high-dimensional dynamical systems in carefully-established low-dimensional latent spaces, so that the low-dimensionality improves computational efficiency, and the accuracy remains under control \cite{antoulas2000survey, benner2015survey, carlberg2015preserving, hesthaven2016certified, noack2011galerkin, noack2011reduced, quarteroni2015reduced}. Despite the wide variety of approaches that can be found in the literature, we can identify two major categories of ROM approaches: \emph{intrusive} and \emph{non-intrusive} methods. Intrusive ROM techniques require access to the full-order solvers, which is often inconvenient in industrial implementations, especially for the cases involving legacy code and/or readily executed software. To overcome this limitation, non-intrusive ROM techniques have been developed to learn low-dimensional representations primarily from data, without accessing the code of FOMs. Such flexibility of non-intrusive ROMs has recently motivated the development of a vast amount of data-driven methods, such as dynamic mode decomposition \cite{schmid2010dynamic, brunton2019data}, reduced-order operator inference \cite{ghattas2021learning, guo2022bayesian, peherstorfer2016data, qian2020lift}, sparse identification of reduced latent dynamics \cite{bakarji2022discovering, brunton2016discovering, champion2019data,conti2023reduced,gao2022bayesian, schaeffer2017learning}, manifold learning using deep auto-encoders \cite{fresca2021comprehensive,fresca2022pod,lee2020model,otto2019linearly}, data-driven approximation of time-integration schemes \cite{zhuang2021model}, Gaussian process modeling for low-dimensional representations \cite{guo2019data, botteghi2022deep}, and kernel flows \cite{owhadi2019kernel}.

However, learning ROMs from noisy data is extremely challenging. While many works in the literature have focused on learning compact representation of high-dimensional data, for example using neural networks (NNs) \cite{lecun2015deep, goodfellow2016deep}, or quantifying uncertainties using, for example, Gaussian processes (GPs) \cite{williams2006gaussian}, herein we argue that both challenges must be tackled simultaneously for discovering ROMs properly. NNs excel in learning complex and nonlinear dependencies of the data, while they tend to struggle when quantifying uncertainties. Conversely, GPs struggle with large datasets due to the limited expressivity of the kernels and the need for the inversion of the covariance matrix, while they excel to quantify uncertainties. To overcome these limitations, and to leverage on the positive features of both NNs and GPs synergistically,  deep kernel learning (DKL) and stochastic variational DKL (SVDKL) were introduced in recent years \cite{wilson2016deep, wilson2016stochastic}. DKL aims to combine the best of the GP and NN worlds  by constructing expressive deep kernels that can model complex relations of the data. DKL builds a deep kernel by feeding to a GP the data processed by a NN.
In addition, SVDKL relies of variational inference \cite{blei2017variational} to allow for the use of traditional minibatch training techniques employed by NN-based models, making SVDKL suitable for effectively dealing with large dataset. Variational inference amortizes the cost of sampling from the (non-Gaussian) posterior distribution by approximating the distribution with the best-fitting Gaussian, reducing the computational cost of SVDKL models.

Another approach to learn expressive deep kernels is kernel flows \cite{owhadi2019kernel, hamzi2021simple, darcy2021learning}. Kernel flows are based on the simple idea that good kernels maintain similar accuracy if the data points are reduced. {\color{black}Kernel flows progressively refine the kernel using subsets of the training set. Thus, they offer a solution to scale GPs to large datasets.} Similarly to DKL, kernel flows can use sequences of nonlinear transformations to pre-process the data before feeding them to the GP kernel. Kernel flows have been successfully used for modeling low-dimensional dynamical systems \cite{yang2024hausdorff, yang2023learning, darcy2023one} even in the case of irregular sampling of the data. Although kernel flows have been applied to low-dimensional dynamical systems, it is worth mentioning that this approach is not inherently limited by the data dimensionality. For example, kernel flows have also been successfully applied to the MNIST dataset \cite{owhadi2019kernel}. 

\begin{figure}[b!] 
    \centering
    \includegraphics[width=0.75\textwidth]{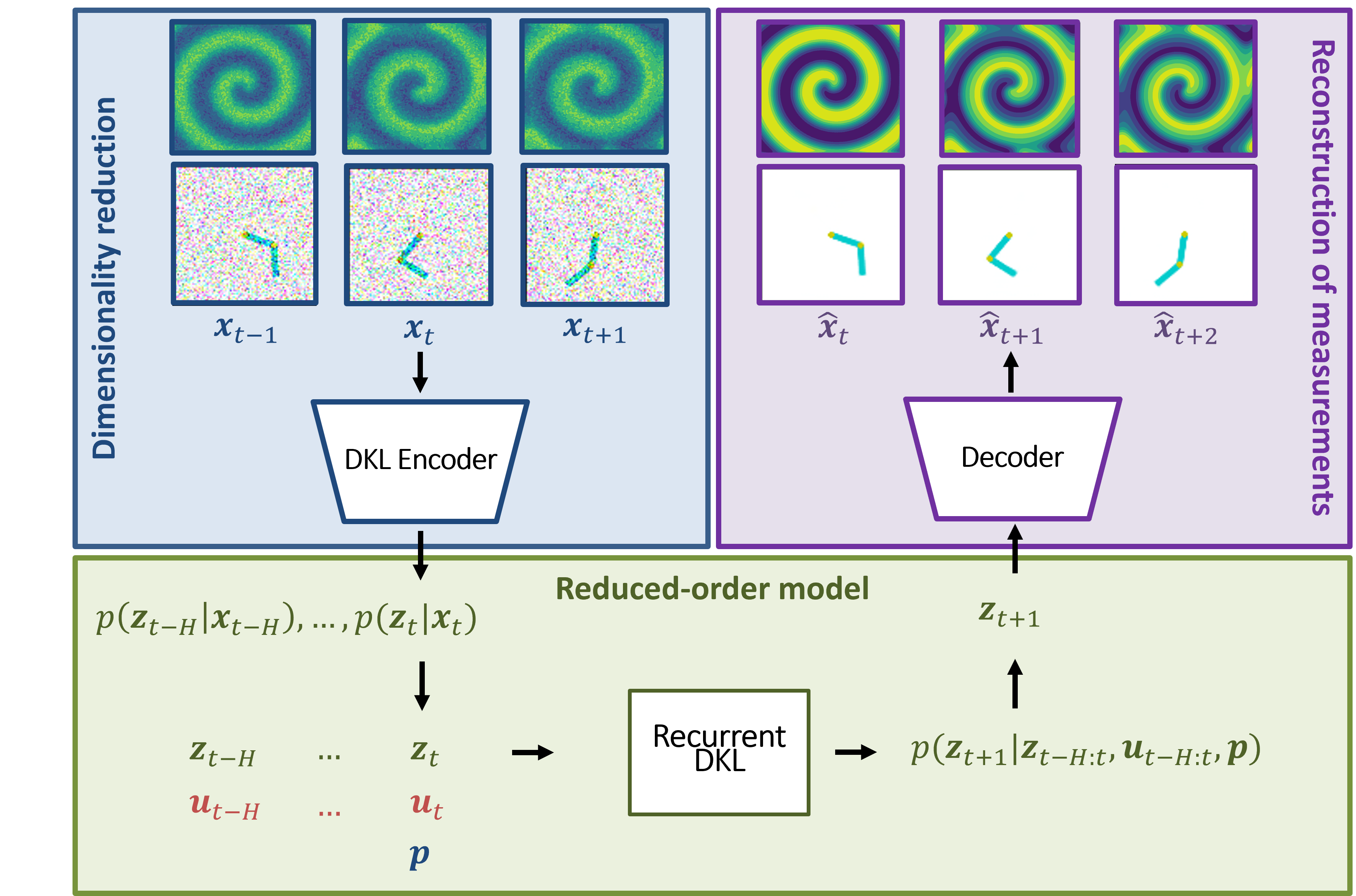}
    \caption{Proposed framework for reduced-order modeling of dynamical systems. We encode the measurements $\mathbf{x}_t$ at different time instances into  the latent variables $\mathbf{z}_t$ by means of a deep kernel learning encoder. Then, we feed a sequence of length $H$ of consecutive latent variables $\mathbf{z}_{t-H:t}$, actions $\mathbf{u}_{t-H:t}$, and parameters $\mathbf{p}$ to a recurrent deep kernel learning to predict the next latent variable $\mathbf{z}_{t+1}$. The measurements are then reconstructed $\hat{\mathbf{x}}$ by means of a decoder from the latent variables. }
    \label{fig:framework}
\end{figure}

In this paper, we exploit the SVDKL idea to develop a non-intrusive ROM that can deal with both high-dimensional and noisy data. In particular, with reference to Figure \ref{fig:framework}, our method includes a SVDKL encoder to compress high-dimensional measurements into low-dimensional distributions of state variables, a recurrent SVDKL latent dynamical model to predict the system's evolution over time, and a decoder to reconstruct the measurements and interpret the latent representations. 
The model is trained without labeled data in an unsupervised manner by only relying on high-dimensional and noisy measurements of the system. We show the capabilities of our framework in two challenging numerical examples, namely, the dynamics of {\em (i)} a double pendulum and of {\em (ii)} a distributed reaction-diffusion system, assuming in both cases to deal with measurements over a two-dimensional spatial region at different time instants -- thus mimicking a set of observations acquired by a camera. Our contribution is threefold:
\begin{itemize}
    \item we improve the prediction accuracy and consistency over long horizons of the SVDKL framework introduced in \cite{botteghi2022deep} by modeling the latent dynamics using a recurrent NN,
    \item we show the capabilities of the framework on challenging chaotic systems, i.e., a double pendulum and a reaction-diffusion problem, and
    \item we introduce an interpretable way for visualizing and studying uncertainties over latent variables by looking at the standard deviation in the measurement space.
\end{itemize}

The paper is organized as follows: in Section \ref{sec:preliminaries}, we introduce the building blocks of our framework, namely GPs, and DKL, and in Section \ref{Sec: Methods}, we describe our novel DKL-based method. Section \ref{sec:numericalresults} presents the numerical experiments, shows and discusses the results, and Section \ref{sec:conclusion} concludes the paper.

\section{Preliminaries}\label{sec:preliminaries}

Throughout this section, we assume to deal with a dataset of $N$ input vectors $X = [ \mathbf{x}_1,..., \mathbf{x}_N ]$, in which  the input is an element $\mathbf{x} \in \mathbb{R}^{|\mathbf{x}|}$ and $|\mathbf{x}|$ denotes  its dimensionality, and correspondingly a vector of targets $\mathbf{y} = [y_1, ..., y_N]^T $, with the output $y \in \mathbb{R}$.

\subsection{Gaussian Processes} \label{subsec:GPs}

 A Gaussian process (GP) \cite{williams2006gaussian} $f(\mathbf{x}) \sim \mathcal{GP}(\mu, k_{\boldsymbol{\gamma}})$ is a collection of random variables, any finite number of which is Gaussian distributed: 
\begin{equation}
    f(\mathbf{x}) \sim \mathcal{GP}(\mu(\mathbf{x}), k(\mathbf{x},\mathbf{x}'; {\boldsymbol{\gamma}})), \ \ \ \ \ {y}=f(\mathbf{x}) + \epsilon, \ \ \ \ \ \epsilon \sim \mathcal{N(}0, \sigma_\epsilon^2) \, .
\end{equation}
A GP is characterized by its mean function $\mu(\mathbf{x}) = \mathbb{E}[f(\mathbf{x})]$ and its covariance/kernel function  $k(\mathbf{x},\mathbf{x}'; {\boldsymbol{\gamma}})=k_{\boldsymbol{\gamma}}(\mathbf{x},\mathbf{x}')=\mathbb{E}[(f(\mathbf{x})-\mu(\mathbf{x}))(f(\mathbf{x}')-\mu(\mathbf{x}'))]$ with hyperparameters $\boldsymbol{\gamma}$, $\mathbf{x}$ and $\mathbf{x}'$ being two input locations, and $\epsilon$ is an additive noise term with variance $\sigma_\epsilon^2$. A popular choice of kernel is the squared-exponential (SE) kernel: 
\begin{equation}
    k_{{\boldsymbol{\gamma}}}(\mathbf{x}, \mathbf{x}') = \sigma_f^2\exp\left(-\frac{1}{2} \frac{||\mathbf{x} - \mathbf{x}'||^2}{l^2}\right) \, ,
\label{eq:RBF_kernel}
\end{equation}
with $\boldsymbol{\gamma}=[\sigma_f^2, l]$, $\sigma_f^2$ being a noise scale and $l$ the length scale. The hyperparameters of the GP regression, namely $[\boldsymbol{\gamma}, \sigma_\epsilon^2]$, can be estimated by maximizing the marginal likelihood as follows:
\begin{equation}
\begin{split}
      [\boldsymbol{\gamma}^*,(\sigma_\epsilon^{2})^*] &= \arg \max_{\boldsymbol{\gamma},\sigma_\epsilon^2}~\log p( \mathbf{y}|X) \\
          &= \arg \max_{\boldsymbol{\gamma},\sigma_\epsilon^2} \left\{-\frac{1}{2} \mathbf{y}^T(k_{\boldsymbol{\gamma}}(X,X)+\sigma_\epsilon^2 I)^{-1}\mathbf{y} -\frac{1}{2} \log |k_{\boldsymbol{\gamma}}(X,X)+\sigma_\epsilon^2I|-\frac{N}{2}\log(2\pi)\right\}\, , \\ 
\end{split}
    \label{eq:marginal_likelihood}
\end{equation}
Given the training data of input-output pairs $(X,\mathbf{y})$, the standard rule for conditioning Gaussians gives a predictive (posterior) Gaussian distribution of the noise-free outputs $\mathbf{f}^*$ at unseen test inputs ${X}^*$:
\begin{equation}
\begin{split}
     \mathbf{f}^*| X^*, X, \mathbf{y} &\sim \mathcal{N}(\boldsymbol{\mu}^*, \Sigma^*)\,,\\
     \boldsymbol{\mu}^*&=k_{\boldsymbol{\gamma}}(X, X^*)^T(k_{\boldsymbol{\gamma}}(X,X)+\sigma_\epsilon^2 I)^{-1}(\mathbf{y}-\mu(X))\,,\\
      \Sigma^*&=k_{\boldsymbol{\gamma}}(X^*, X^*) - k_{\boldsymbol{\gamma}}(X, X^*)^T(k_{\boldsymbol{\gamma}}(X,X)+\sigma_\epsilon^2 I)^{-1}k_{\boldsymbol{\gamma}}(X, X^*)\,. \\
\end{split}
\end{equation}

\subsection{Deep Kernel Learning}

The choice of the kernel is a crucial aspect for the performance of a GP. For example, a SE kernel (see Equation \ref{eq:RBF_kernel}) can only learn information about the data using the noise-scale and the length-scale parameters, which tell us how quickly the correlation in our data varies with respect to the distance of the input-data pairs. This is clearly a limiting factor when dealing with complex datasets with non-trivial dependencies and correlations. To overcome the problem of the limited expressivity of GP kernels, deep kernel learning (DKL) was introduced in \cite{williams2006gaussian}. 

The key idea of DKL is to embed a (deep) NN $g(\mathbf{x};\boldsymbol{\theta})$ of parameters $\boldsymbol{\theta}$, i.e., the weights and biases of the NN, into the kernel function of a GP:
 \begin{equation}
     k_{\text{DKL}}(\mathbf{x}, \mathbf{x'}; {\boldsymbol{\gamma}}, \boldsymbol{\theta}) = k_{{\boldsymbol{\gamma}}}(g(\mathbf{x}; \boldsymbol{\theta}), g(\mathbf{x'}; \boldsymbol{\theta}))=k(g(\mathbf{x}; \boldsymbol{\theta}), g(\mathbf{x'}; \boldsymbol{\theta});\boldsymbol{\gamma})\, .
 \end{equation}
In this way, we can rely on the expressive power of NNs to learn compact and low-dimensional representations of the data, and unveil the non-trivial dependencies of the data. Then, we can feed these representations to the GP for quantifying the uncertainties. 

The GP hyperparameters $\boldsymbol{\gamma}$ and the NN parameters $\boldsymbol{\theta}$ are jointly learned  by maximizing the log marginal likelihood (see Equation \eqref{eq:marginal_likelihood}). It is possible to use the chain rule to compute derivatives of the log marginal likelihood, that we indicate with $\mathcal{L}(\boldsymbol{\gamma}, \boldsymbol{\theta})$, with respect to kernel hyperparameters $\boldsymbol{\gamma}$ and the NN parameters $\boldsymbol{\theta}$, thus obtaining:
\begin{equation}
    \frac{\partial \mathcal{L}(\boldsymbol{\gamma}, \boldsymbol{\theta})}{\partial {\boldsymbol{\gamma}} } = \frac{\partial \mathcal{L}(\boldsymbol{\gamma}, \boldsymbol{\theta})}{\partial k_{{\boldsymbol{\gamma}}}} \frac{\partial k_{{\boldsymbol{\gamma}}}}{\partial {\boldsymbol{\gamma}}},  \hspace{0.7cm} \frac{\partial \mathcal{L}(\boldsymbol{\gamma}, \boldsymbol{\theta})}{\partial \boldsymbol{\theta}} = \frac{\partial \mathcal{L}(\boldsymbol{\gamma}, \boldsymbol{\theta})}{\partial k_{{\boldsymbol{\gamma}}}} \frac{\partial k_{{\boldsymbol{\gamma}}}}{\partial g(\mathbf{x}, \boldsymbol{\theta})} \frac{\partial g(\mathbf{x}, \boldsymbol{\theta})}{\partial \boldsymbol{\theta}}.
\end{equation}
 Optimizing the GP hyperparameters through Equation \eqref{eq:marginal_likelihood} requires to repeatedly invert the covariance matrix $k_{\boldsymbol{\gamma}}(X,X)+\sigma_\epsilon^2I$, with size $N \times N$. However, for large datasets, i.e., when $N \gg 1$, the full covariance matrix may be too large to be stored in memory and computationally prohibitive to invert it. In these cases, it is common practice in machine learning to utilize minibatches of randomly-samples data points to compute the loss functions and compute the gradients. However, when we use minibatches to train the GPs, the posterior distribution is not Gaussian anymore, even when a Gaussian likelihood is used. When dealing with non-Gaussian posteriors, we can rely on variational inference (VI) \cite{blei2017variational}. VI is an approximation technique for dealing with non-Gaussian posteriors, deriving from minibatch training and/or non-Gaussian likelihoods. VI allows one to approximate the posterior by the best-fitting Gaussian distribution using a set of samples, often called inducing points, from the posterior. VI allows for drastically reducing the computation cost of standard GPs and effectively utilizing DKL with large datasets.

Stochastic variational DKL (SVDKL) \cite{wilson2016stochastic} extends DKL to minibatch training and non-Gaussian posteriors by means of VI. SVDKL can deal with large dataset, effectively mitigating the main limitations of traditional GPs. SVDKL is also considerably cheaper than Bayesian NNs or ensembles methods \cite{goan2020bayesian, rosen1996ensemble}, making it an essential architecture in many applications. SVDKL can be used as a building block for learning ROMs from high-dimensional and noisy data \cite{botteghi2022deep}. The framework proposed in \cite{botteghi2022deep} is composed of (i) a SVDKL encoder-decoder scheme for learning low-dimensional representations of the data and (ii) a SVDKL using the representations to predict the dynamics of the systems forward in time. The encoder-decoder scheme resembles a variational autoencoder \cite{kingma2013auto}, but with better uncertainty quantification capabilities due to the presence of the GP kernel. 

\section{Recurrent Stochastic Variational Deep Kernel Learning for Dynamical Systems}\label{Sec: Methods}

\subsection{Problem Statement}
\label{Subsection: Problem Statement}
In our research, we consider nonlinear dynamical systems expressed in the state-space form:
\begin{equation}
   \dot{\mathbf{s}}(t) = f(\mathbf{s}(t), \mathbf{u}(t); \mathbf{p}),  \hspace{0.5 cm} \mathbf{s}(t_0) = \mathbf{s}_0, \hspace{0.5 cm}  t \in [t_0,t_f]\, ,
\end{equation}
where $\mathbf{s}(t) \in \mathbb{R}^{|\mathbf{s}|}$ represents the state vector at time $t$, $\dot{\mathbf{s}}(t)$ its time derivative, $\mathbf{u}(t) \in \mathbb{R}^{|\mathbf{u}|}$ is the control input at time $t$, $\mathbf{p} \in \mathbb{R}^{|\mathbf{p}|}$ is the vector of parameters, $\mathbf{s}_0$ is the initial condition, and $t_0$ and $t_f$ are the initial and final times, respectively. The nonlinear function $f$ determines the evolution of the system with respect to the current state $\mathbf{s}(t)$ and control input $\mathbf{u}(t)$.

In many real-world applications, the state $\mathbf{s}(t)$ and the FOM $f$ may be unknown or not readily available. However, we can often obtain indirect information about these systems through measurements collected by sensor devices. We denote the measurements at a generic timestep $t$ with $\mathbf{x}_t$, with $\mathbf{x} \in \mathbb{R}^{|\mathbf{x}|}$, and the measurement at timestep $t+1$ as $\mathbf{x}_{t+1}$ due to the discrete-time nature of the measurements. Given a set of $M$ observed trajectories $\Xi = [X_1, \cdots, X_M]$,  controls $\Omega = [U_1, \cdots ,U_M]$,  and parameters $P=[\mathbf{p}_1, \cdots, \mathbf{p}_{M} ]$, if any, where for each trajectory we collect $N$ high-dimensional and noisy measurements $X_i=[\mathbf{x}_1, \cdots, \mathbf{x}_N]$, $N-1$ control inputs $U_i=[\mathbf{u}_1, \cdots, \mathbf{u}_{N-1}]$, and one parameter vector $\mathbf{p}_i$, our objective is to introduce a framework for: (i) learning a compact representation $\mathbf{z}$ of the unknown state variables $\mathbf{s}$, and (ii) learning a surrogate ROM $\xi$ as a proxy for $f$ that predicts the dynamics of the latent state variables, given, if any, control inputs and parameters. Due to the high dimensionality and corruption of the measurement data, and the unsupervised nature of the learning process\footnote{We do not rely on the true environment states in our framework.}, achieving objectives (i) and (ii) is extremely challenging.  

For conciseness, we use simplified notation $\mathbf{x}_t^{-H}$ to represent the series $\mathbf{x}_{t-H}, \cdots, \mathbf{x}_t$ throughout the paper, with $H$ denoting the history length. This applies not only to the full states $\mathbf{x}$ but also to the latent states $\mathbf{z}$ and control inputs $\mathbf{u}$.

\subsection{Model Architecture}\label{Subsec: Model_Architecture}

\begin{figure}[h!]
    \centering
    \includegraphics[width=0.5\textwidth]{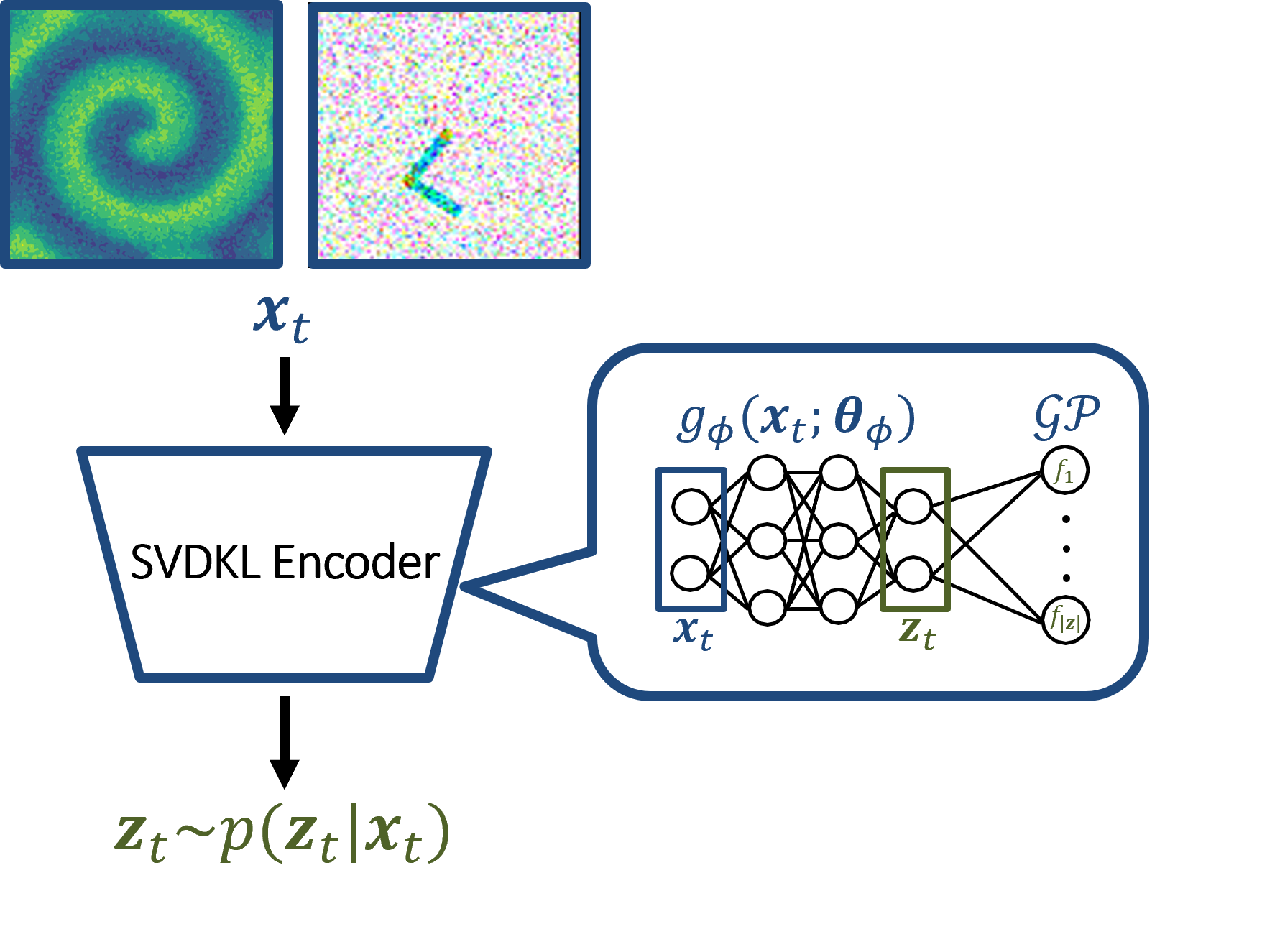}
    \caption{Architecture of the SVDKL encoder.}
    \label{fig:1a}
\end{figure}

\begin{figure}[h!]
    \centering
    \includegraphics[width=0.5\textwidth]{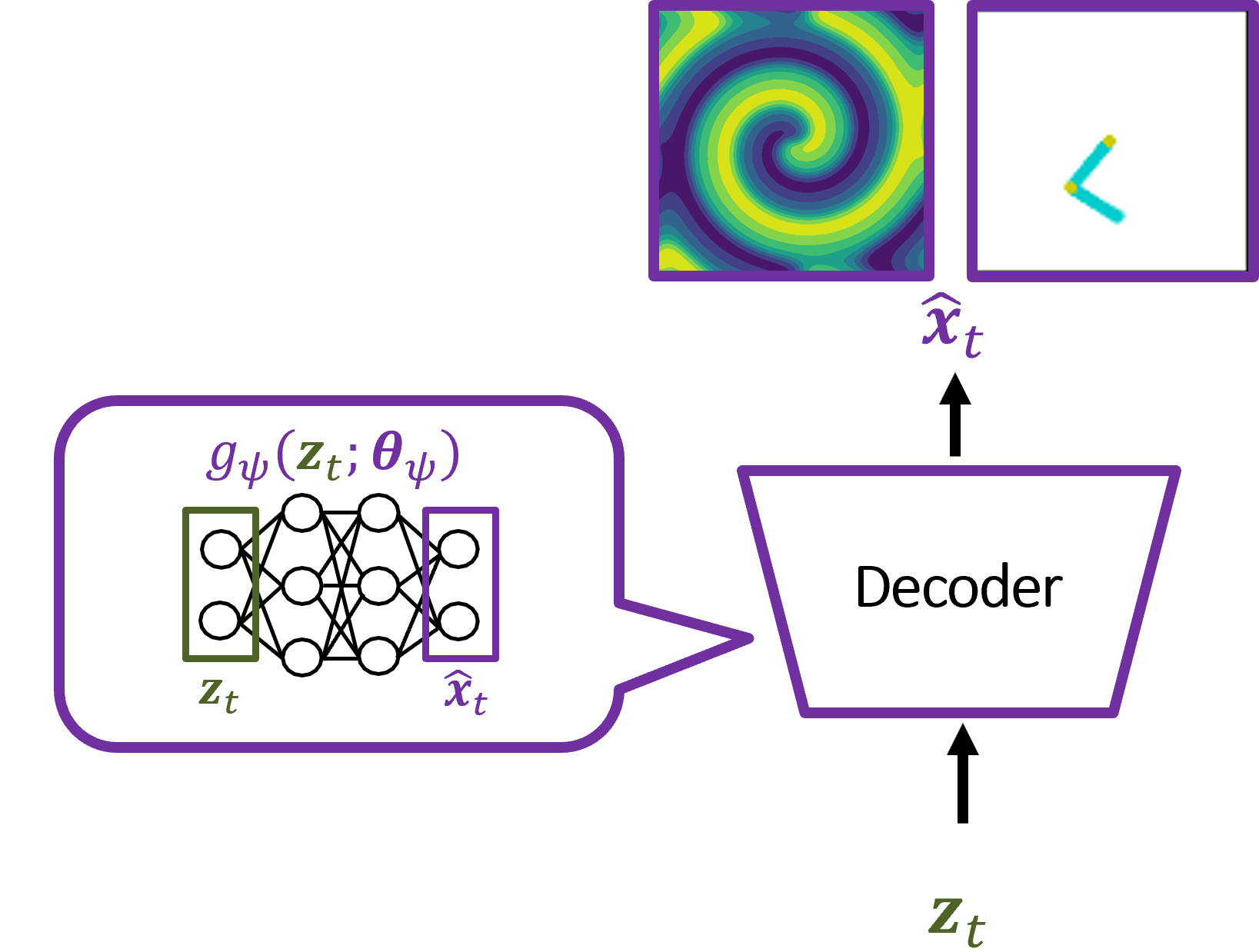}
    \caption{Architecture of the decoder.}
    \label{fig:1b}
\end{figure}

\begin{figure}[h!]
    \centering
    \includegraphics[width=0.8\textwidth]{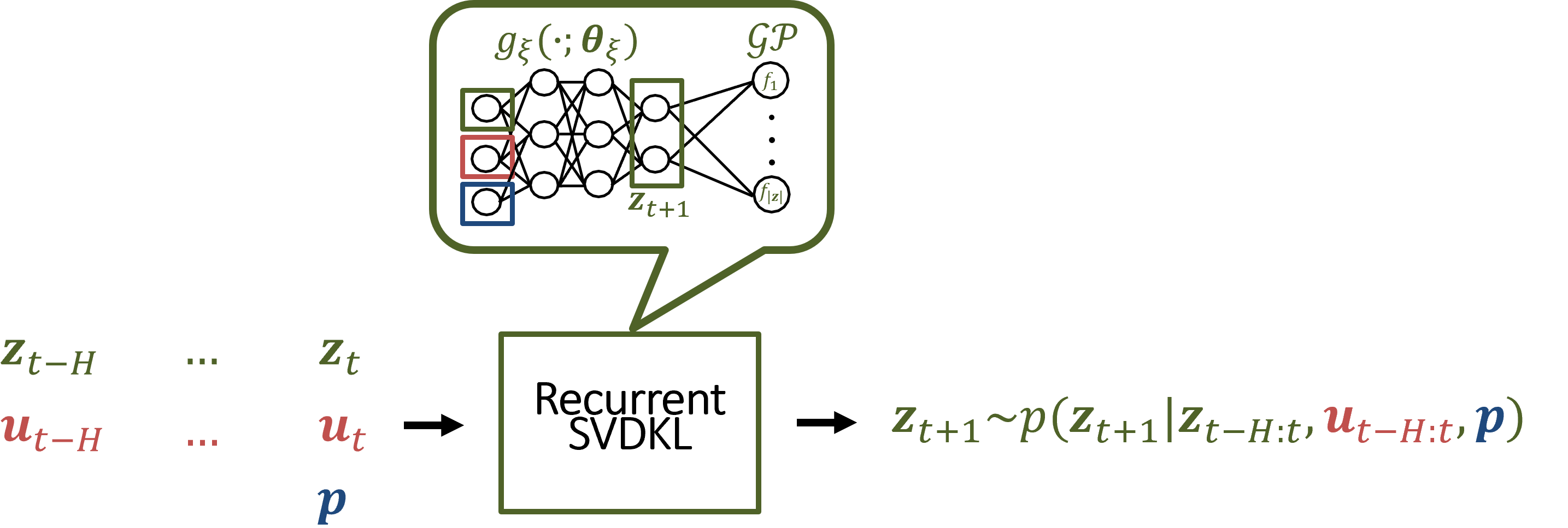}
    \caption{Architecture of the forward dynamical model.}
    \label{fig:1c}
\end{figure}

To tackle these two challenges, we introduce a novel recurrent SVDKL architecture (see Figure \ref{fig:framework}) that is composed of three main blocks (see Figure \ref{fig:1a}, \ref{fig:1b}, and \ref{fig:1c}, respectively): 
\begin{enumerate}
    \item an SVDKL encoder that projects the high-dimensional measurements $\mathbf{x}$ into a Gaussian distribution over the latent state variables $p(\mathbf{z}|\mathbf{x})$,
    \item a decoder that reconstructs the measurements $\hat{\mathbf{x}}$ from the latent variables $\mathbf{z}$, and
    \item a recurrent SVDKL forward dynamical model predicting the evolution of the system using a history of latent state variables $\mathbf{z}_{t}^{-H}$ and control inputs  $\mathbf{u}_{t}^{-H}$, and parameters $\mathbf{p}$, if available. 
\end{enumerate}

With reference to Figure \ref{fig:1a}, the encoder $\phi: \mathbb{R}^{|\mathbf{x}|}\rightarrow [0, 1]^{|\mathbf{z}|}$ is modeled using a SVDKL of parameters $\boldsymbol{\theta}_{\phi}$ and hyperparameters $\boldsymbol{\gamma}_{\phi}, \sigma_{\phi}^2$, with $[0, 1]^{|\mathbf{z}|}$ indicating the distribution over the latent variables $\mathbf{z}$. In particular, the encoder $\phi$ maps a high-dimensional measurement $\mathbf{x}_t$ to a latent state distribution $p(\mathbf{z}_t|\mathbf{x}_t)$:
\begin{equation}
\begin{split}
 z_{i,t} &= f_i^{\phi}(\mathbf{x}_t) + \epsilon_{\phi},\quad \epsilon_{\phi} \sim \mathcal{N}(0, \sigma^2_{\phi})\, ,\\
     f_i^{\phi}(\mathbf{x}_t) &\sim \mathcal{GP}(\mu(g_{\phi}(\mathbf{x}_t; \boldsymbol{\theta}_{\phi})), k(g_{\phi}(\mathbf{x}_t; \boldsymbol{\theta}_{\phi}), g_{\phi}(\mathbf{x}'_{t'}; \boldsymbol{\theta}_{\phi});\boldsymbol{\gamma}_{\phi,i})),\quad 1\leq i \leq |\mathbf{z}|\,,
\end{split}
\label{eq:posteriorE}
\end{equation}
where $z_{i,t}$ indicates the sample from the $i^{th}$ GP with SE kernel $k(\bullet, \bullet';\boldsymbol{\gamma}_{\phi})$ and mean $\mu$, $\epsilon_{\phi}$ is an independently-added noise, and $|\mathbf{z}|$ indicates the dimension of $\mathbf{z}$. The GP inputs $g_{\phi}(\mathbf{x}_t; \boldsymbol{\theta}_{\phi})$ and $g_{\phi}(\mathbf{x}'_t; \boldsymbol{\theta}_{\phi})$ are the representations of the data pair $(\mathbf{x}, \mathbf{x}')$ obtained from the NN $g_{\phi}(\bullet;\boldsymbol{\theta}_{\phi})$ with parameters $\boldsymbol{\theta}_{\phi}$. 

With reference to Figure \ref{fig:1b}, the decoder $\psi$ is modeled using an NN with parameters $\boldsymbol{\theta}_{\psi}$. The decoder maps a sample $\mathbf{z}$ to the conditional distribution $p(\hat{\mathbf{x}}|\mathbf{z})$. Similar to VAEs \cite{kingma2013auto}, the distribution $p(\hat{\mathbf{x}}|\mathbf{z})$ is approximated to be independently-distributed Gaussian distributions. In general, the decoding NN $\psi$ learns the means and variances of these distributions. Following \cite{hafner2019learning, hafner2019dream}, however, our decoder in this work only learns the mean vector of $\hat{\mathbf{x}}|\mathbf{z}$, while the covariance is set to the identity matrix $I$, i.e., $\psi: \mathbb{R}^{|\mathbf{z}|}\rightarrow [0,1]^{|\mathbf{x}|}$ and
\begin{equation}
           \hat{\mathbf{x}}_t|\mathbf{z}_t \sim \mathcal{N}\psi(\mathbf{z}_t; \boldsymbol{\theta}_{\psi}), I).
\label{eq:dec}
\end{equation}
The reconstruction $\hat{\mathbf{x}}_t$ can be obtained by feeding $\mathbf{z}_t$ to the decoder $\psi$ and then sampling from this distribution.

With reference to Figure \ref{fig:1c}, the forward dynamical model  $\xi: \mathbb{R}^{|\mathbf{z}|}\times  \mathbb{R}^{|\mathbf{u}|} \times  \mathbb{R}^{|\mathbf{p}|}\rightarrow [0, 1]^{|\mathbf{z}|}$, with parameters $\boldsymbol{\theta}_{\xi}$ and hyperparameters $\boldsymbol{\gamma}_{\xi}, \sigma_{\xi}^2$, is a SVDKL with a recurrent NN architecture, i.e., a long short-term memory (LSTM) NN \cite{graves2012long}, that maps sequences of latent states, actions, and parameters to latent next state distribution $p\left(\mathbf{z}_{t+1} | \mathbf{z}_{t}^{-H},\mathbf{u}_{t}^{-H},\mathbf{p}\right)$.

In particular, we can write the latent next states $\mathbf{z}_{t+1}$ as:
\begin{equation}
\begin{split}
z_{i,t+1} &= f_i^{\xi}\left(\mathbf{z}_{t}^{-H},\mathbf{u}_{t}^{-H}, \mathbf{p}\right) + \epsilon_{\xi},\quad \epsilon_{\xi} \sim \mathcal{N}(0, \sigma^2_{\xi})\,, \\
f_i^{\xi}(\bullet) &\sim \mathcal{GP}(\mu(g_{\xi}(\bullet;\boldsymbol{\theta}_{\xi}), k(g_{\xi}(\bullet; \boldsymbol{\theta}_{\xi}), g_\xi(\bullet'; \boldsymbol{\theta}_{\xi});\boldsymbol{\gamma}_{{\xi},i})) \,,\quad 1\leq i \leq |\mathbf{z}|\,,
\end{split}
\label{eq:posteriorF}  
\end{equation}
where $z_{i,t+1}$ is sampled from the $i^{th}$ GP, $H$ is the history length, and $\epsilon_{\xi}$ is a noise term. Similarly to the encoder, the GP inputs $g_{\xi}\left(\mathbf{z}_{t}^{-H},\mathbf{u}_{t}^{-H}, \mathbf{p}; \boldsymbol{\theta}_{\xi}\right)$ and $g_{\xi}\left((\mathbf{z}')_{t}^{-H},(\mathbf{u}')_{t}^{-H}, \mathbf{p}'; \boldsymbol{\theta}_{\xi}\right)$ are the representations of obtained from the NN $g_{\xi}(\bullet;\boldsymbol{\theta}_{\xi})$ with parameters $\boldsymbol{\theta}_{\xi}$. 

The encoder and decoder have the same NN architecture proposed in \cite{botteghi2022deep}. On the other hand, the forward dynamical model replaces the fully-connected NN used in \cite{botteghi2022deep} with a recurrent neural network to improve the reliability and accuracy of the predictions forward in time. 

\subsection{Training Objectives}

The aforementioned encoder, decoder, and forward dynamical model are jointly trained to infer meaningful low-dimensional representations of the measurements and predict the system evolution forward in time. 
The first objective term is formulated by utilizing the reconstruction loss, commonly used for training VAEs, which maximizes the marginal likelihood\footnote{In practice this is achieved by minimizing the negative log-marginal likelihood.} for the reconstruction of ${\mathbf{x}}_t$ by $\hat{\mathbf{x}}_t$:
\begin{equation}
    \mathcal{L}_{\text{recon}}(\boldsymbol{\theta}_{\phi},\boldsymbol{\gamma}_{\phi}, \sigma^2_{\phi}, \boldsymbol{\theta}_{\psi}) = \mathbb{E}_{\mathbf{x}_t \sim \Xi}[-\log \mathbb{E}_{{\mathbf{z}}_t|\mathbf{x}_t} p(\hat{\mathbf{x}}_t|\mathbf{z}_t)]\,,
\label{eq:loss_SVDKL_VAE}
\end{equation}
where $\hat{\mathbf{x}}_t$ is the reconstructed vector of the measurement $\mathbf{x}_t$, obtained by feeding $\mathbf{x}_t$ through the encoder $\phi$ and decoder $\psi$, $p(\mathbf{z}|\mathbf{x})$ is the distribution learned by the encoder (see Equation \eqref{eq:posteriorE}), and $p(\hat{\mathbf{x}}|\mathbf{z})$ the distribution learned by the decoder (see Equation \eqref{eq:dec}). The minimization of loss function in Equation (\ref{eq:loss_SVDKL_VAE}) with respect to the encoder and decoder parameters is analogous to the one commonly used by VAEs, which allows us to learn compact representations of the measurements while quantifying the uncertainties in the reconstructions.

For training the dynamical model $\xi$, we do not rely on true state values $\mathbf{s}$, but on the sequences of measurements collected at different time steps $t$. Therefore, we are not able to use supervised learning techniques to optimize the NN parameters and kernel hyperparameters.
However, we can use the posterior distribution $p(\mathbf{z}_{t+1} | \mathbf{x}_{t+1})$ (see Equation \eqref{eq:posteriorE}) as the target distribution for learning $p\left(\mathbf{z}_{t+1} | \mathbf{z}_t^{-H},\mathbf{u}_t^{-H},\mathbf{p}\right)$. 
This process translates into evaluating the Kullback-Leiber ($\text{KL}$) divergence between these two distributions:
\begin{equation}
\begin{split}
    & \mathcal{R}_{\text{reg}}^{(1)}(\boldsymbol{\theta}_{\phi},\boldsymbol{\gamma}_{\phi}, \sigma^2_{\phi}, \boldsymbol{\theta}_{\psi},\boldsymbol{\theta}_{\xi},\boldsymbol{\gamma}_{\xi}, \sigma^2_{\xi})\\
    =~ & \text{KL}\left[p(\mathbf{z}_{t+1} | \mathbf{x}_{t+1}) \Big\|~ \mathbb{E}_{\mathbf{z}_t^{-H}|\mathbf{x}_{t}^{-H}}~p\left(\mathbf{z}_{t+1} | \mathbf{z}_t^{-H},\mathbf{u}_t^{-H},\mathbf{p}\right)  \right] \, ,
\end{split}
\label{eq:loss_SVDKL_VAE_onestep}
\end{equation}
where the superscript $(1)$ in $\mathcal{R}_{\text{reg}}^{(1)}$ indicates that we perform a 1-step ahead prediction of the latent states from time step $t$ to $t+1$ and the subscript reg stands for regularization. The distribution $p(\mathbf{z}_{t+1} | \mathbf{x}_{t+1})$ is obtained by encoding the next measurement $\mathbf{x}_{t+1}$ through $\phi$ (see Figure \ref{fig:1a}), $p\left(\mathbf{z}_{t+1} | \mathbf{z}_{t}^{-H},\mathbf{u}_{t}^{-H},\mathbf{p}\right)$ is the distribution obtained by the forward dynamical model $\xi$ (see Figure \ref{fig:1c}), and the expectation $\mathbb{E}_{\mathbf{z}_t^{-H}|\mathbf{x}_t^{-H}}$ indicates the marginalization of a sequence of latent variables $\mathbf{z}_t^{-H}$ obtained by feeding the measurements $\mathbf{x}_t^{-H}$ to the encoder $\phi$.
Different from \cite{botteghi2022deep}, we extend this loss to $T$-step predictions:
\begin{equation}
    \mathcal{L}_{\text{reg}}(\boldsymbol{\theta}_{\phi},\boldsymbol{\gamma}_{\phi}, \sigma^2_{\phi}, \boldsymbol{\theta}_{\psi},\boldsymbol{\theta}_{\xi},\boldsymbol{\gamma}_{\xi}, \sigma^2_{\xi}) = \mathbb{E}_{\mathbf{x} \sim \Xi, \mathbf{u} \sim \Omega,  \mathbf{p}\sim P} \frac{1}{T}\sum_{i=1}^T \mathcal{R}_{\text{reg}}^{(i)}(\boldsymbol{\theta}_{\phi},\boldsymbol{\gamma}_{\phi}, \sigma^2_{\phi}, \boldsymbol{\theta}_{\psi},\boldsymbol{\theta}_{\xi},\boldsymbol{\gamma}_{\xi}, \sigma^2_{\xi})\, ,
    \label{Eq:Loss_F_multistep_reg}
\end{equation}
where we now perform $T$-step ahead predictions from time step $t+1$ to $t+T$, and 
\begin{equation}
\begin{split}
    & \mathcal{R}_{\text{reg}}^{(i)}(\boldsymbol{\theta}_{\phi},\boldsymbol{\gamma}_{\phi}, \sigma^2_{\phi}, \boldsymbol{\theta}_{\psi},\boldsymbol{\theta}_{\xi},\boldsymbol{\gamma}_{\xi}, \sigma^2_{\xi})\\
    =~ & \text{KL}\left[p(\mathbf{z}_{t+i} | \mathbf{x}_{t+i}) \Big\|~ \mathbb{E}_{\mathbf{z}_{t+i-1}^{-H} \big| \mathbf{x}_{t+i-1}^{-H}}p\left(\mathbf{z}_{t+i} ~|~ \mathbf{z}_{t+i-1}^{-H},\mathbf{u}_{t+i-1}^{-H},\mathbf{p}\right)  \right].
\end{split}
\label{eq:loss_SVDKL_VAE_R}
\end{equation}
Similar to Equation \eqref{eq:loss_SVDKL_VAE_onestep}, the expectation $\mathbb{E}_{\mathbf{z}_{t+i-1}^{-H}|\mathbf{x}_{t+i-1}^{-H}}$ indicates the marginalization of  a sequence of latent variables $\mathbf{z}_{t+i-1}^{-H}$ obtained by feeding the measurements $\mathbf{x}_{t+i-1}^{-H}$ to the encoder $\phi$ (Figure \ref{fig:1a}).

Additionally, we include a $T$-step reconstruction loss of the ``next measurements'' for timesteps $t+1, \cdots, t+T$, which is a variant of the reconstruction loss in Equation \eqref{eq:loss_SVDKL_VAE} to optimize the parameters in the encoder, decoder, and forward model:
\begin{equation}
    \mathcal{L}_{\text{recon-next}}(\boldsymbol{\theta}_{\phi},\boldsymbol{\gamma}_{\phi}, \sigma^2_{\phi}, \boldsymbol{\theta}_{\psi},\boldsymbol{\theta}_{\xi},\boldsymbol{\gamma}_{\xi}, \sigma^2_{\xi}) = \mathbb{E}_{\mathbf{x} \sim \Xi, \mathbf{u} \sim \Omega,  \mathbf{p}\sim P} \frac{1}{T}\sum_{i=1}^T \mathcal{R}_{\text{recon-next}}^{(i)}(\bullet)\, ,
    \label{Eq:Loss_F_multistep}
\end{equation}
where the superscript $(i)$ indicated the $i^{\text{th}}$ step's reconstruction loss
\begin{equation}
\mathcal{R}_{\text{recon-next}}^{(i)} (\boldsymbol{\theta}_{\phi},\boldsymbol{\gamma}_{\phi}, \sigma^2_{\phi}, \boldsymbol{\theta}_{\psi},\boldsymbol{\theta}_{\xi},\boldsymbol{\gamma}_{\xi}, \sigma^2_{\xi}) = \left\|\mathbf{x}_{t+i}-\mathbb{E}[\hat{\mathbf{x}}_{t+i} ~|~ \mathbf{x}_{t+i-1}^{-H},\mathbf{u}_{t+i-1}^{-H},\mathbf{p} ]\right\|^2\, ,
\end{equation}
and the marginalized $\hat{\mathbf{x}}_{t+i}$ in the expectation $\mathbb{E}\left[\hat{\mathbf{x}}_{t+i} |\mathbf{x}_{t+i-1}^{-H}\right]$ is sampled by encoding the sequence of measurements $\mathbf{x}_{t+i-1}^{-H}$ through $\phi$ (see Figure \ref{fig:1a}), predicting the next latent states through $\xi$ (see Figure \ref{fig:1c}), and then decoding through $\psi$ (see Figure \ref{fig:1b}):
\begin{equation}
 \mathbb{E}\left[\hat{\mathbf{x}}_{t+i} ~|~\mathbf{x}_{t+i-1}^{-H},\mathbf{u}_{t+i-1}^{-H},\mathbf{p} \right]
= \mathbb{E}_{\mathbf{z}_{t+i-1}^{-H}|\mathbf{x}_{t+i-1}^{-H}} \mathbb{E}_{\mathbf{z}_{t+i} | \mathbf{z}_{t+i-1}^{-H},\mathbf{u}_{t+i-1}^{-H},\mathbf{p}} \mathbb{E}\left[ \hat{\mathbf{x}}_{t+i}|\mathbf{z}_{t+i} \right].
\end{equation}

Eventually, the two SVDKL models, $\phi$ and $\xi$, utilize VI to approximate the posterior distributions in Equations \eqref{eq:posteriorE} and \eqref{eq:posteriorF}, respectively. Thus, in addition to the aforementioned loss terms, we include the VI losses $\text{KL}[p(\mathbf{v})|| q(\mathbf{v})]$ for both SVDKL models. Here $p(\mathbf{v})$ is the posterior to be approximated over the inducing points $\mathbf{v}$, and $q(\mathbf{v})$ represents an approximating candidate distribution. The total VI loss term is the sum of the encoder's VI loss $\mathcal{L}_{\text{VI}}^{\phi}(\boldsymbol{\gamma}_{\phi}, \sigma_{\phi}^2)$ and the forward dynamical model's VI loss $\mathcal{L}_{\text{VI}}^{\xi}(\boldsymbol{\gamma}_{\xi}, \sigma_{\xi}^2)$:
\begin{equation}
    \mathcal{L}_{\text{VI}} = \mathcal{L}_{\text{VI}}^{\phi}(\boldsymbol{\gamma}_{\phi}, \sigma_{\phi}^2) + \mathcal{L}_{\text{VI}}^{\xi}(\boldsymbol{\gamma}_{\xi}, \sigma_{\xi}^2)\,.
    \label{Eq: VL_loss}
\end{equation}

The overall loss function that will be jointly optimized is simply the weighted sum of all the aforementioned objective terms:
\begin{equation}
    \mathcal{L}_{\text{total}}(\boldsymbol{\theta}_{\phi},\boldsymbol{\gamma}_{\phi}, \sigma^2_{\phi}, \boldsymbol{\theta}_{\psi}, \boldsymbol{\theta}_{\xi},\boldsymbol{\gamma}_{\xi}, \sigma^2_{\xi}) = \mathcal{L}_{\text{recon}} + \omega_{\text{reg}}\mathcal{L}_{\text{reg}} + \mathcal{L}_{\text{recon-next}} +  \omega_{\text{var}}\mathcal{L}_{\text{VI}}\, ,
\label{eq:total_loss}
\end{equation}
in which $\omega_{\text{reg}}$ and $\omega_{\text{var}}$ are scalar factors chosen to balance the contribution of the different terms. We perform a simple grid search to find these coefficients and leave more advanced strategies for hyperparameter optimization to future work.

\section{Numerical Results}\label{sec:numericalresults}
\begin{figure}[b!]
    \centering
    \includegraphics[width=0.5\textwidth]{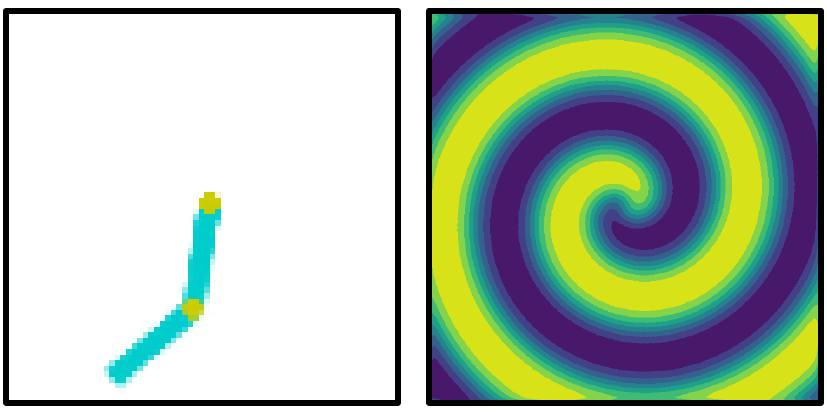}
    \caption{Double pendulum and reaction-diffusion systems.}
    \label{fig: numerical_examples}
\end{figure}

We test our framework on two commonly-studied, yet complex, baselines: (i) a double pendulum with an actuated joint, and (ii) a parametric reaction-diffusion system (see Figure \ref{fig: numerical_examples}). Both systems present chaotic dynamics, making their evolution over time hard to capture accurately.
 We assume to have access only to high-dimensional and noisy measurements $\mathbf{x}$ of these systems. In our experiments, we aim to (a) denoise of the measurements, (b) learn compact and interpretable representations, (c) predict the  systems' dynamics, and (d) quantify uncertainties.

\textbf{Denoising of Measurements}. When dealing with noisy data, an aspect of paramount importance is the ability of the models to remove the noise from the data, that is, denoising. To test the denoising capabilities of our framework, we add Gaussian noise over the measurements:
\begin{equation}
\begin{split}
        \mathbf{x}_t &= \mathbf{x}_t + \boldsymbol{\epsilon}_{\mathbf{x}} \\
        \boldsymbol{\epsilon}_{\mathbf{x}} &\sim \mathcal{N}(\mathbf{0}, \sigma^2_{\mathbf{x}}I)\, , \\
\end{split}
\end{equation}
where $\sigma^2_{\mathbf{x}} \in \{0.0, 0.25^2, 0.5^2\}$ corresponds to the variance. We aim at recovering the original measurements (see Figure \ref{fig: Multi-step SVDKL Denoising T20} and \ref{fig: Multi-step SVDKL Denoising T20 PDE}). In addition, to provide a quantitative evaluation, we utilize the peak signal-to-noise ratio (PSNR) and the $L_1$ norm to assess the denosing abilities of our method. The PSNR measures the ratio between the maximum possible power of a signal, i.e., $\max(\mathbf{x}^2_t)$, and the power of the corrupting noise, i.e., $|| \mathbf{x}_t -  \hat{\mathbf{x}}_t ||_2^2$. The PSNR is defined as:
\begin{equation}
\text{PSNR} = 10 \cdot \log_{10}\left(\frac{\max (\mathbf{x}_t^2)}{|| \mathbf{x}_t -  \hat{\mathbf{x}}_t ||_2^2}\right)\, ,
\end{equation}
where $\mathbf{x}_t$ and $\hat{\mathbf{x}}_t$ correspond to the (noisy) measurement and its reconstruction at a generic timestep $t$, respectively. The $L_1$ norm measures how close the recovered measurement $\hat{\mathbf{x}}_t$ is to the uncorrupted measurement $\mathbf{x}_t$. The $L_1$ norm is the absolute difference between the original and reconstructed measurements:
\begin{equation}
L_1 = ||\mathbf{x}_t - \hat{\mathbf{x}}_t||_1 = |\mathbf{x}_t - \hat{\mathbf{x}}_t|.
\end{equation}
We report these results in Table \ref{tab:quant_results_double_pendulum} and \ref{tab:quant_results_PDE}.

\textbf{Learning Compact and Interpretable Representations}. The second aspect we are interested in is assessing the ability of the framework to learn compact and interpretable latent representations of the system states. To visualize the latent (state) representations, we employ the t-distributed stochastic neighbor embedding (t-SNE) method \cite{van2008visualizing} to project the latent states to a 2-dimensional space and inspect their correlation with the true state variables (see Figure \ref{fig: Multi-step SVDKL Denoising T20} and \ref{fig: Multi-step SVDKL Denoising T20 PDE}).

\textbf{Predicting the Dynamics}. The goal of our method is to learn a ROM that can accurately and efficiently predict the evolution of different dynamical systems. Therefore, after training the model, we evaluate its predictions forward in time by feeding the initial state to the ROM and autoregressively predict the trajectory of the systems. We then compare the predicted trajectories with the trajectories from a test set that were not used for the training of the model (see Figure \ref{fig: Multi-step SVDKL Predicting Dynamics T10} and \ref{fig: Multi-step SVDKL Predicting Dynamics T10 PDE}.

\textbf{Quantifying Uncertainties}. Ultimately, we would like to quantify uncertainties, deriving from the noisy measurements, properly. However, uncertainties over the latent variables are hard to visualize and, consequently analyze \cite{botteghi2022deep}. Therefore, we study uncertainties in the measurement space that can be more informative then uncertainties over the latent variables. We generate multiple rollouts of the ROM for a fixed initial condition, we decode the latent-space trajectories back to the measurement space using the decoder, we compute standard deviation of the different trajectories, and we plot heatmaps representing the evolution of the uncertainties over time (see Figure \ref{fig:UQ_IC_det_F_sampling_T=20} and \ref{fig:UQ_IC_det_F_sampling_T=20_0.0_PDE}). 

\subsection{Double Pendulum}
\label{Subsection: Equation of Motion}
The first example we consider is an actuated double pendulum. The double pendulum exhibits chaotic and nonlinear behavior, making the prediction of its dynamics from high-dimensional and noisy measurements extremely challenging. Its dynamics can be described as a function of its joint angles   $(\theta_1, \theta_2)$, velocities $(\dot{\theta}_1, \dot{\theta}_2)$, and accelerations $(\ddot{\theta}_1, \ddot{\theta}_2)$. We indicate with  $\theta_1, \theta_2$ the angle of the first joint and second joint, $l_1, l_2$ the length of the two links, and $m_1, m_2$ the mass of the two links. The equations of motions of the double pendulum can be written as in state-space form as:
\begin{equation}
    \begin{split}
        \dot{\theta}_1 &= \omega_1 \,,\\
        \dot{\theta}_2 &= \omega_2 \,,\\
        \dot{\omega}_1 &= \frac{-g(2m_1+m_2)\sin \theta_1 - m_2g\sin(\theta_1 -2 \theta_2)-2\sin(\theta_1 - \theta_2)m_2(\omega_2^2 l_2 + \omega_1^2 l_1 \cos(\theta_1-\theta_2))}{l_1(2m_1+m_2-m_2\cos(2\theta_1 - 2\theta_2))} + u_1\,,\\
        \dot{\omega}_2 &= \frac{2 \sin(\theta_1-\theta_2) (\omega_1^2 l_1 (m_1 + m_2) + g(m_1 + m_2) \cos \theta_1 + \omega_2^2 l_2 m_2\cos(\theta_1-\theta_2))}{l_2 (2 m_1 + m_2 - m_2  \cos(2\theta_1-2\theta_2))}\, , \\
    \end{split}
\label{eq:doublependulum}
\end{equation}
where $\omega_1$ and $\omega_2$ are the angular velocities of the first and second link of the double pendulum, respectively, and $u_1$ the control input (torque) to the first joint. The measurements $\mathbf{x}$ are high-dimensional snapshots of dimension $84 \times 84 \times 3$ of the double pendulum dynamics (see Equation \eqref{eq:doublependulum}) when a random control $u_1$ is applied. An example of measurement is shown in Figure \ref{fig: numerical_examples} (left).

\begin{figure}[h!]
    \centering
    \includegraphics[width=\textwidth]{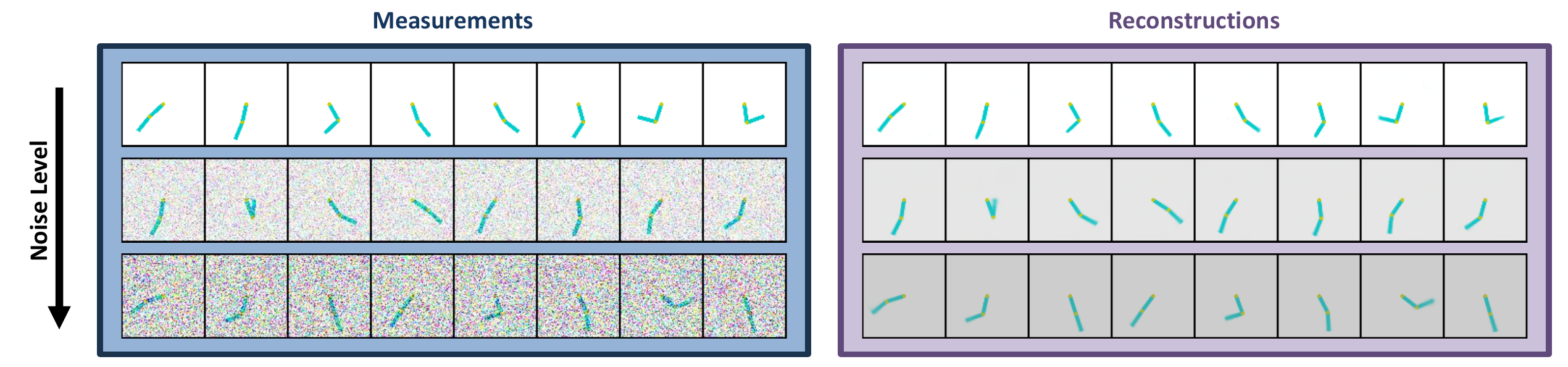}
      \caption{Reconstructions of  $\hat{\mathbf{x}}_t$  with noise levels $\sigma^2_{\mathbf{x}} \in \{0, 0.25^2, 0.5^2\}$ applied to input measurements. The noise level increases from the top to the bottom images.}
    \label{fig: Multi-step SVDKL Denoising T20}
\end{figure}

In Figure \ref{fig: Multi-step SVDKL Denoising T20}, we show the reconstructions $\hat{\mathbf{x}}_t$ and $\hat{\mathbf{x}}_{t+1}$ for different levels of noise of the measurements $\sigma^2_{\mathbf{x}} \in \{0, 0.25^2, 0.5^2\}$. In this case, we set the latent dimension $|\mathbf{z}|=20$ and the history length $H=20$. As shown by the results, the model can properly denoise the noisy measurements, even in the case of high levels of noise ($\sigma^2_{\mathbf{x}}=0.5^2$).The ability to remove the noise from the measurements derives from the ability of the model to encode the relevant features into the latent state.

To further understand the effect of the recurrent SVDKL architecture on the prediction performance of the ROM, we quantitatively analyze, using the PSNR and L$_1$ metric , the quality of the reconstructions $\hat{\mathbf{x}}_t$ and $\hat{\mathbf{x}}_{t+1}$. The results for different values of $H$, where $H=1$ corresponds to the original architecture proposed in \cite{botteghi2022deep}, and noise variance $\sigma^2_{\mathbf{x}}$ are reported in Table \ref{tab:quant_results_double_pendulum}. Increasing the history length $H$ improves the encoding and latent model performance and consequently denoising abilities of the model, as shown by a higher PSNR and a lower $L_1$ norm. However, due to the sequential nature of the LSTM, longer input sequences requires more computations and may slow down the training. We found that a value of $H \in [10, 20]$ provides a good trade-off between results and computational burden.
\begin{table}[h!]
\centering
\begin{tabular}{ |c|@{}c|@{}c|@{}| }
\hline
 Noise Level & Reconstruction $\hat{\mathbf{x}}_t$ & Reconstruction $\hat{\mathbf{x}}_{t+1}$ \\ [0.5ex] 
\hline \hline
$\sigma^2_{\mathbf{x}}=0.0$ & 
\begin{tabular}{cccc} 
$H$ & $T$ & \text{PSNR (db)} & \text{L}$_1$ \\ [0.25ex] 
\hline
1 & 3 &  29.12 & 53.74 \\
10 & 3 & 33.20 & 22.25\\
20 & 3 &  31.89 & 22.71\\ 
\end{tabular} &
\begin{tabular}{cccc}
$H$ & $T$ & \text{PSNR (db)} & \text{L}$_1$ \\ [0.25ex] 
\hline
1 & 3 &  21.72 & 208.00 \\
10 & 3 & 31.33 & 32.18\\
20 & 3 &  30.21 & 37.50\\ 
\end{tabular}\\

\hline
$\sigma^2_{\mathbf{x}}=0.25^2$ & 
\begin{tabular}{cccc}
$H$ & $T$ & \text{PSNR (db)} & \text{L}$_1$ \\ [0.25ex] 
\hline
1 & 3 &  26.97 & 86.15\\
10 & 3 & 29.26 & 62.39\\
20 & 3 &  29.45 & 60.14\\ 
\end{tabular} &
\begin{tabular}{cccc}
$H$ & $T$ & \text{PSNR (db)} & \text{L}$_1$ \\ [0.25ex] 
\hline
1 & 3 &  22.98 & 180.30\\
10 & 3 & 28.19 & 74.69\\
20 & 3 &  28.44 & 72.55\\ 
\end{tabular}\\
\hline

$\sigma^2_{\mathbf{x}}=0.5^2$ & 
\begin{tabular}{cccc}
$H$ & $T$ & \text{PSNR (db)} & \text{L}$_1$ \\ [0.25ex] 
\hline
1 & 3 &  23.59 & 179.97\\
10 & 3 & 24.58 & 155.60\\
20 & 3 &  24.77 & 150.84\\ 
\end{tabular} &
\begin{tabular}{cccc}
$H$ & $T$ & \text{PSNR (db)} & \text{L}$_1$ \\ [0.25ex] 
\hline
1 & 3 &  20.23 & 322.15\\
10 & 3 & 24.46 & 159.01\\
20 & 3 & 24.59 & 156.12\\ 
\end{tabular}\\
\hline
\end{tabular}
\caption{Quantitative results in the case of the double pendulum for different values of $H$. All the models are trained using the loss function in Equation \eqref{eq:total_loss} with $T=3$.}
\label{tab:quant_results_double_pendulum}
\end{table}

\begin{figure}[h!]
    \centering
    \includegraphics[width=\textwidth]{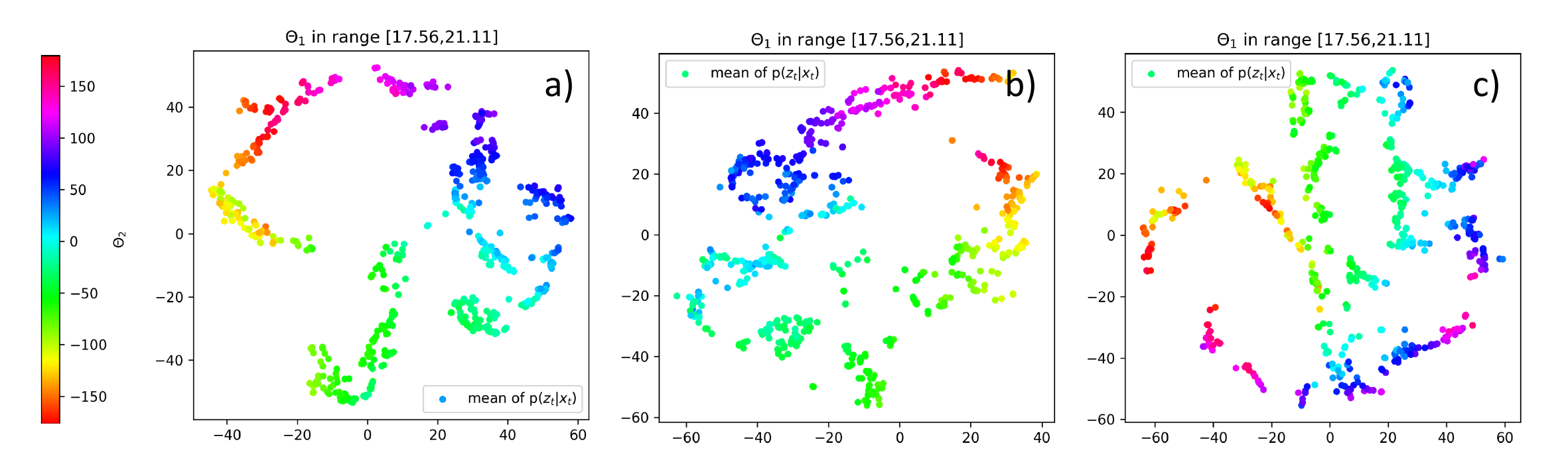}
    \caption{t-SNE visualization of the mean of the latent state distribution for different levels of noise $\sigma^2_{\mathbf{x}} \in \{0, 0.25^2, 0.5^2\}$ on the measurements, Figure a), b), and c), respectively. The first angle $\theta_1$ is fixed within a range, while the color bar represents the second angle $\theta_2$.}
    \label{fig: Multi-step_SVDKL_Latent_Representations_N_20_T20}
\end{figure}

In Figure \ref{fig: Multi-step_SVDKL_Latent_Representations_N_20_T20}, we show the latent variables for different value of noise applied to the measurements. Due to the strong denoising capabilities of our model, the latent representations, visualized via t-SNE, are minimally affected  and their correlation with the true variables of the systems, i.e. the angles $\theta_1$ and $\theta_2$, remains high. Again, we show the results for $|\mathbf{z}|=20$ and $H=20$.

\begin{figure}[h!]
    \centering
    \includegraphics[width=\textwidth]{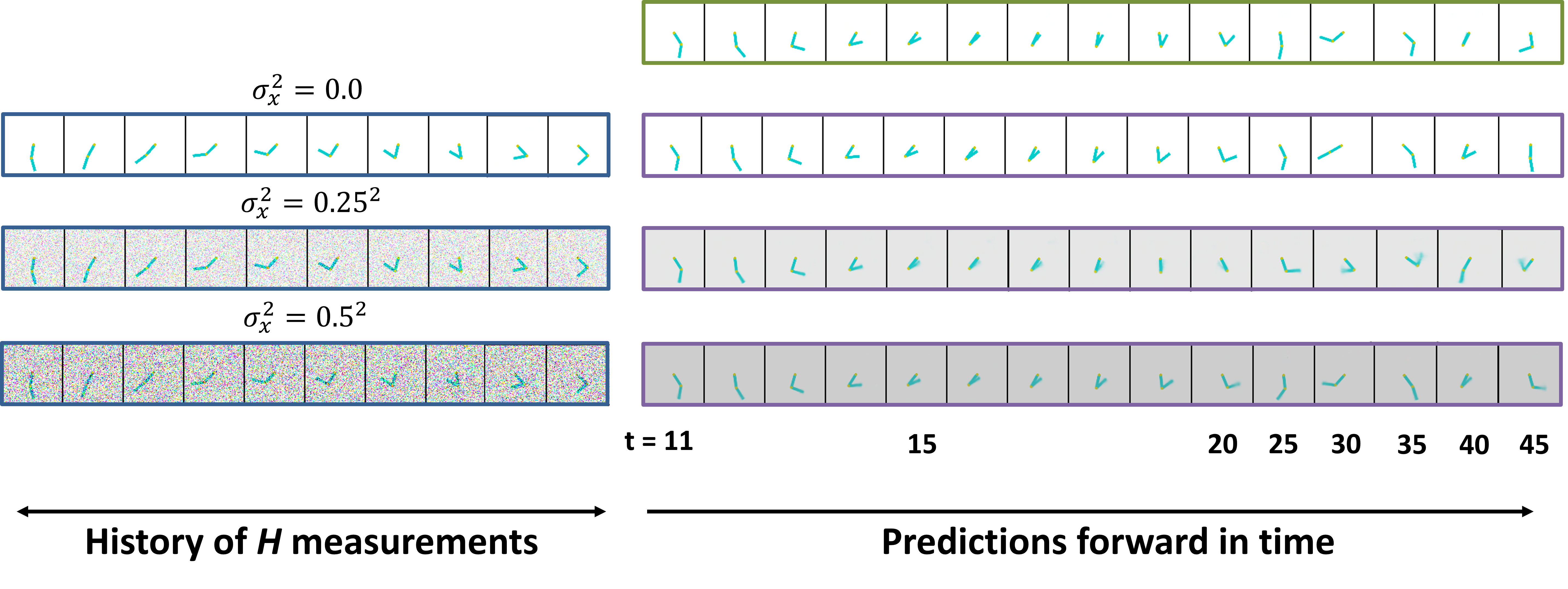}
  \caption{Predictions of the dynamic $\hat{\mathbf{x}}_t$ (purple box) for different noise levels $\sigma^2_{\mathbf{x}} \in \{0, 0.25^2, 0.5^2\}$ applied to input measurements (blue box). The true trajectory is highlighted by the green box.}
  \label{fig: Multi-step SVDKL Predicting Dynamics T10}
\end{figure}

To assess the prediction capability of the proposed ROM, we predict the dynamics forward in time from a history of measurements of length $H$. In Figure \ref{fig: Multi-step SVDKL Predicting Dynamics T10}, we show the reconstruction of the predicted trajectories in comparison with the true ones from the test set.
The reconstructions remain close to the true measurements, even with the highest noise level ($\sigma^2_{\mathbf{x}}=0.5^2$) for the first 20 timesteps. Afterwards, especially in the noisy-measurement cases, we notice a gradual divergence. However, it is worth mentioning that the double pendulum is a chaotic systems and small perturbations of the initial conditions generate drastically different trajectories. Therefore, it is natural to observe this behavior, especially with additive noise on the measurements acting as a perturbation of the initial conditions.

\begin{figure}[h!]
    \centering
    \includegraphics[width=\textwidth]{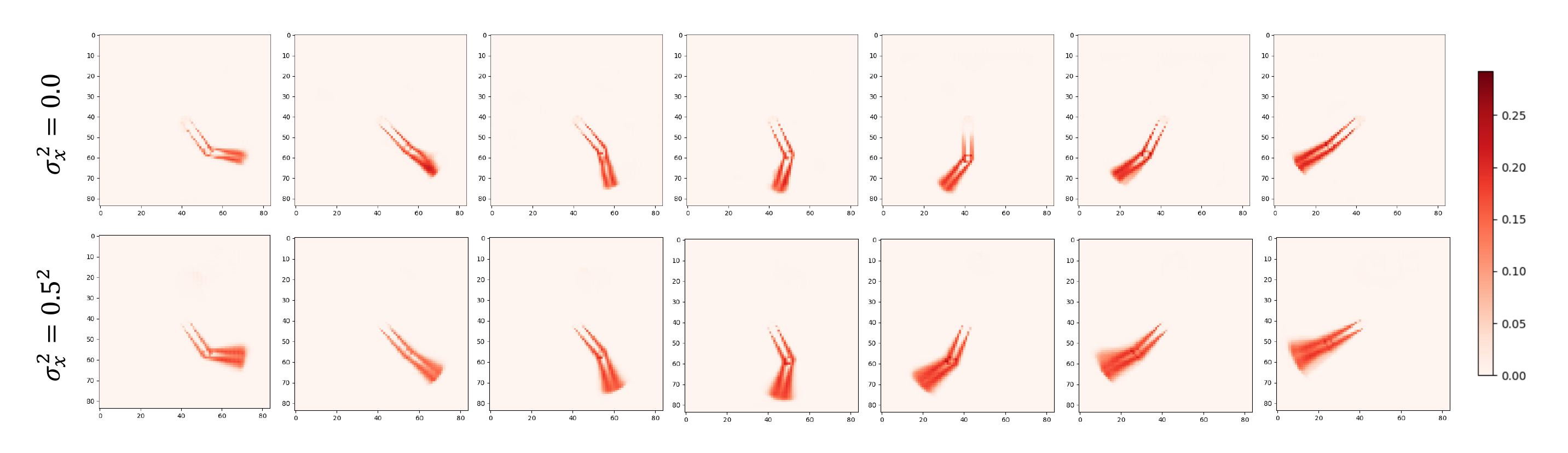}
        \caption{Evolution over time of the standard deviation of the trajectories projected in the image space for $\sigma^2_{\mathbf{x}}=0.0$ and $\sigma^2_{\mathbf{x}}=0.5^2$. }
        \label{fig:UQ_IC_det_F_sampling_T=20}
\end{figure}

In Figure \ref{fig:UQ_IC_det_F_sampling_T=20}, we show the uncertainty quantification capabilities of our framework by analyzing the uncertainties over the same trajectory with different noise levels ($\sigma^2_{\mathbf{x}}=0.0$ and $\sigma^2_{\mathbf{x}}=0.5^2$). 
It is possible to notice that the model is sensitive to the noise added to the measurements, as the standard deviation of the predicted trajectory grows with more noise, i.e., the pendulum position is more blurred in the case of $\sigma^2_{\mathbf{x}}=0.5^2$ than in the case of $\sigma^2_{\mathbf{x}}=0.0$.

\subsection{Nonlinear Reaction-Diffusion Problem} 
The second example we consider is a lambda–omega reaction–diffusion system that can be used to describe a wide variety of physical phenomena, spanning from chemistry to biology and geology. The equations describing the dynamics can be written as:
\begin{equation}
    \begin{split}
        \dot{u} &= (1-(u^2 + v^2))u + \beta(u^2 + v^2)v + d(u_{xx}, u_{yy}) \,,\\
        \dot{v} &= \beta(u^2 + v^2)u + (1-(u^2 + v^2))v  + d(v_{xx}, v_{yy})\, ,\\
    \end{split}
\end{equation}
where $u=u(x,y,t)$ and $v=v(x,y,t)$ describe the evolution of the spiral wave over time in the spatial domain $(x,y) \in [-10, 10]$, and $\beta$ and $d$ are the parameters regulating the reaction and diffusion behavior of the system, respectively. Our parameter of interest is $\beta$, varying in the range $[0.5, 1.5]$. Similarly to \cite{conti2023multi}, we assume periodic boundary conditions and initial condition equal to:
\begin{equation}
    u(x, y, 0) = v(x, y, 0) = \tanh (\sqrt{x^2 + y^2} \cos( (x + iy) - \sqrt{x^2 + y^2}).
\end{equation}
In this case, the measurements $\mathbf{x}$ are obtained by spatially discretizing the PDE with a $128 \times 128$ grid.

\begin{figure}[h!]
    \centering
    \includegraphics[width=\textwidth]{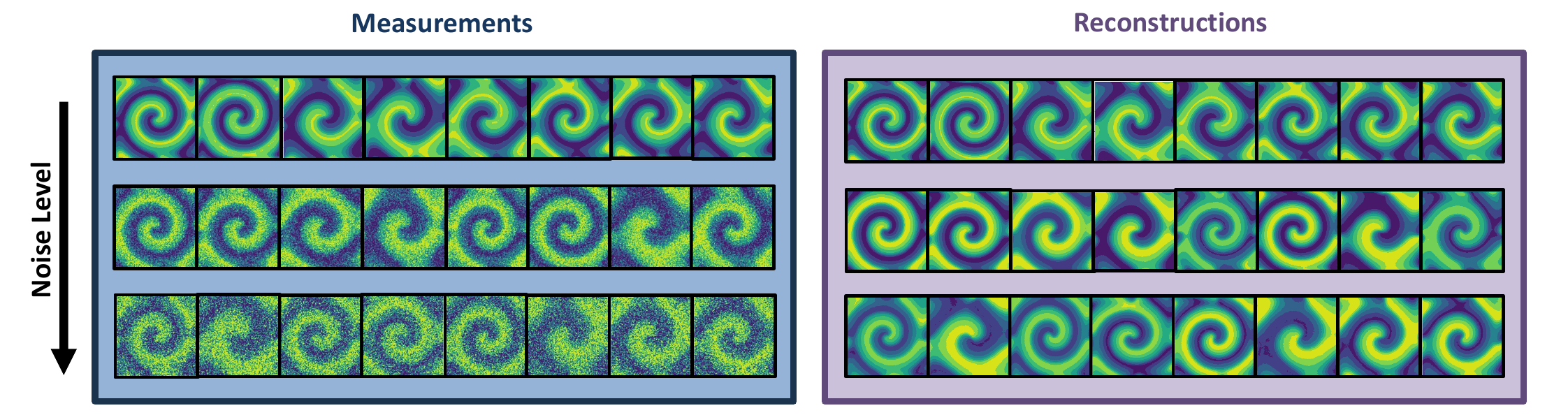}
      \caption{Reconstructions of  $\hat{\mathbf{x}}_t$  with noise levels $\sigma^2_{\mathbf{x}} \in \{0, 0.25^2, 0.5^2\}$ applied to input measurements. The noise level increases from the top to the bottom images.}
  \label{fig: Multi-step SVDKL Denoising T20 PDE}
\end{figure}

In Figure \ref{fig: Multi-step SVDKL Denoising T20 PDE}, we show the reconstructions $\hat{\mathbf{x}}_t$ and $\hat{\mathbf{x}}_{t+1}$ for different levels of noise of the measurements $\sigma^2_{\mathbf{x}} \in \{0, 0.25^2, 0.5^2\}$. Similarly to the double pendulum example, we set the latent dimension $|\mathbf{z}|=20$ and the history length $H=20$. As shown in Figure \ref{fig: Multi-step SVDKL Denoising T20 PDE}, the model can properly denoise the noisy measurements, even in the case of high levels of noise ($\sigma^2_{\mathbf{x}}=0.5^2$). 

Similarly to the case of the double pendulum,  we quantitatively analyze, using the PSNR and L$_1$ metrics, the quality of the reconstructions $\hat{\mathbf{x}}_t$ and $\hat{\mathbf{x}}_{t+1}$. The results for different values of $H$, where $H=1$ corresponds to the original architecture proposed in \cite{botteghi2022deep}, and noise $\sigma^2_{\mathbf{x}}$ are reported in Table \ref{tab:quant_results_PDE}. Even in the reaction-diffusion PDE, increasing the history length $H$ improves the encoding and latent model performance and consequently denoising abilities of the model.
\begin{table}[h!]
\centering
\begin{tabular}{ |c|@{}c|@{}c|@{}| }
\hline 
 & State reconstruction $\hat{\mathbf{x}}_t$ & Next state reconstruction $\hat{\mathbf{x}}_{t+1}$ \\
\hline \hline
$\sigma_{\mathbf{x}}^2=0.0$ & 
    \begin{tabular}{cccc}
$H$ & $T$ & \text{PSNR (db)} & \text{L}$_1$ \\
\hline 
1 & 3 &  29.32 & 347.91 \\
10 & 3 & 29.29 & 360.26\\
20 & 3 &  29.34 & 342.73\\ 
\end{tabular} &
\begin{tabular}{cccc}
$H$ & $T$ & \text{PSNR (db)} & \text{L}$_1$ \\
\hline
1 & 3 &  29.30 & 352.20 \\
10 & 3 & 29.05 & 401.94\\
20 & 3 &  29.31 & 344.89\\ 
\end{tabular}\\

\hline
$\sigma_{\mathbf{x}}^2=0.25^2$ & 
\begin{tabular}{cccc}
$H$ & $T$ & \text{PSNR (db)} & \text{L}$_1$ \\
\hline
1 & 3 &  26.55 & 1029.88\\
10 & 3 & 26.52 & 1039.02\\
20 & 3 &  26.50 & 1042.06\\ 
\end{tabular} &
\begin{tabular}{cccc}
$H$ & $T$ & \text{PSNR (db)} & \text{L}$_1$ \\
\hline
1 & 3 &  26.56 & 1024.53\\
10 & 3 & 26.50 & 1040.52\\
20 & 3 &  26.54 & 1030.01\\ 
\end{tabular}\\
\hline

$\sigma_{\mathbf{x}}^2=0.5^2$ & 
\begin{tabular}{cccc}
$H$ & $T$ & \text{PSNR (db)} & \text{L}$_1$ \\
\hline
1 & 3 &  19.82 & 2947.03\\
10 & 3 & 19.83 & 2940.93\\
20 & 3 &  19.85 & 2945.56\\ 
\end{tabular} &
\begin{tabular}{cccc}
$H$ & $T$ & \text{PSNR (db)} & \text{L}$_1$ \\
\hline
1 & 3 &  19.88 & 2923.69\\
10 & 3 & 19.84 & 2939.25\\
20 & 3 & 19.82 & 2943.92\\ 
\end{tabular}\\
\hline
\end{tabular}
\caption{Quantitative results in the case of the reaction-diffusion PDE for different values of $H$. All the models are trained using the loss function in Equation \eqref{eq:total_loss} with $T=3$.}
\label{tab:quant_results_PDE}
\end{table}

\begin{figure}[h!]
    \centering
    \includegraphics[width=\textwidth]{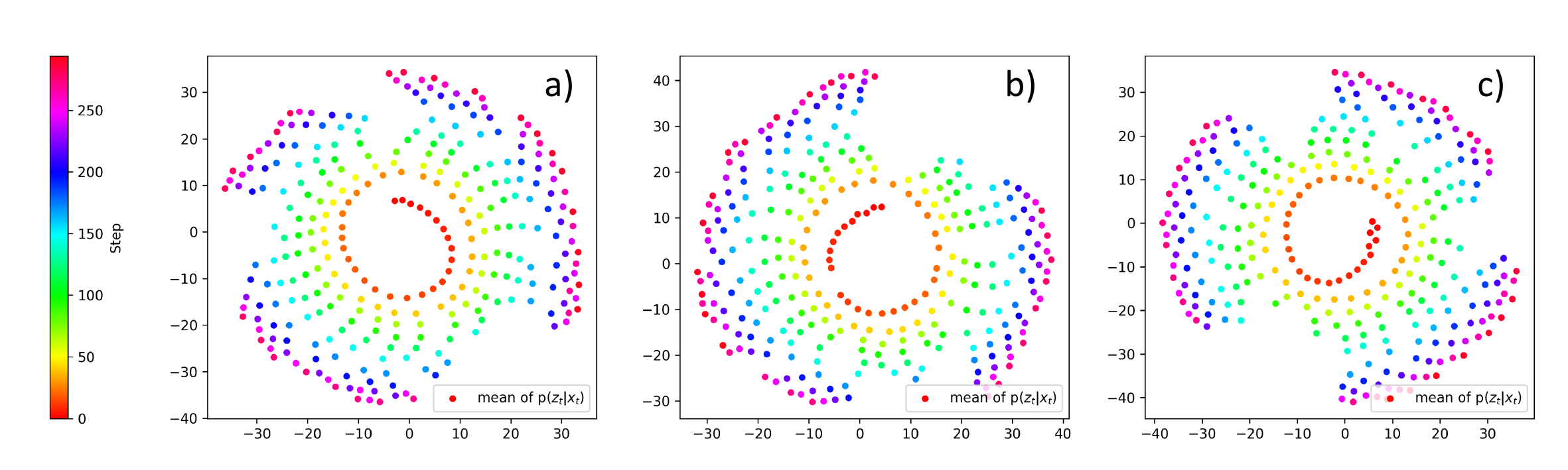}
    \caption{t-SNE visualization of the mean of the latent state distribution for different levels of noise $\sigma^2_{\mathbf{x}} \in \{0, 0.25^2, 0.5^2\}$ on the measurements, Figure a), b), and c), respectively. The colorbar represents the timesteps.}
    \label{fig: Multi-step_SVDKL_Latent_Representations_N_20_T20_DE}
\end{figure}

In Figure \ref{fig: Multi-step_SVDKL_Latent_Representations_N_20_T20_DE}, we show the latent variables for different value of noise applied to the measurements. Similarly to the pendulum example, due to the strong denoising capabilities of our model, the latent representations, visualized via t-SNE, are minimally affected  by the noise. This aspect can be noticed by the fact that the latent representations do no qualitatively change for different levels of noise. Again, we show the results for $|\mathbf{z}|=20$ and $H=20$.

\begin{figure}[h!]
    \centering 
    \includegraphics[width=1.0\textwidth, angle=0]{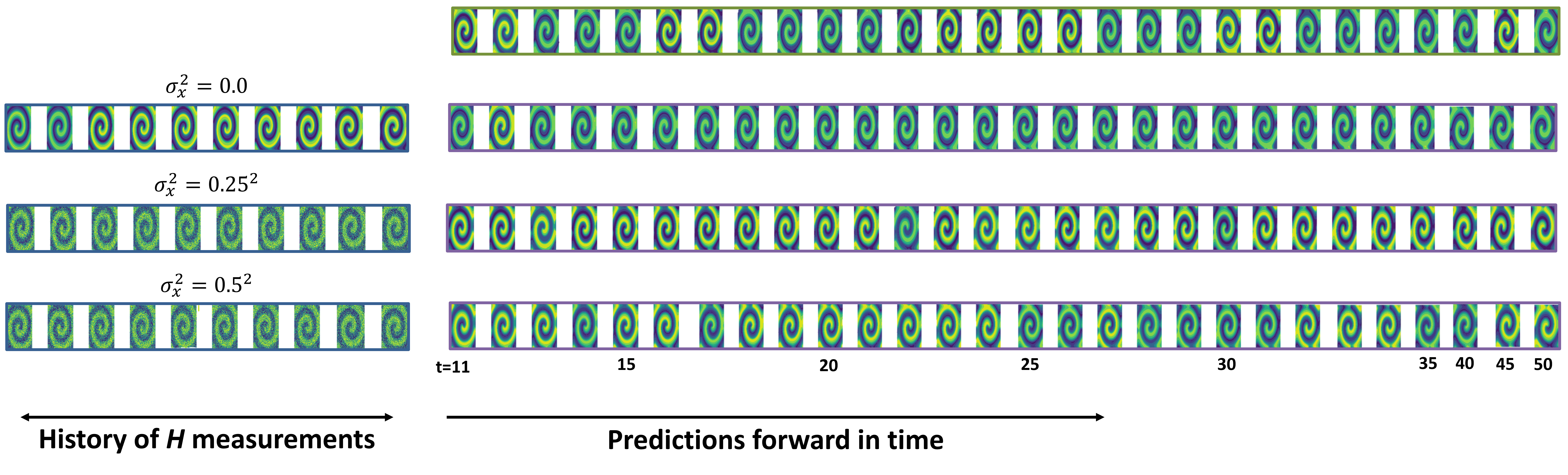}
  \caption{Predictions of the dynamic $\hat{\mathbf{x}}_t$ (purple box) for different noise levels $\sigma^2_{\mathbf{x}} \in \{0, 0.25^2, 0.5^2\}$ applied to input measurements (blue box). The true trajectory is highlighted by the green box.}
  \label{fig: Multi-step SVDKL Predicting Dynamics T10 PDE}
\end{figure}

To assess the prediction capability of the proposed ROM, we predict the dynamics forward in time from a history of measurements of length $H$. In Figure \ref{fig: Multi-step SVDKL Predicting Dynamics T10 PDE}, we show the reconstruction of the predicted trajectories in comparison with the true ones from the test set.
The reconstructions remain close to the true measurements, even with the highest noise level ($\sigma_\mathbf{x}^2=0.5^2$).

\begin{figure}[h!]
    \centering
    \includegraphics[width=\textwidth]{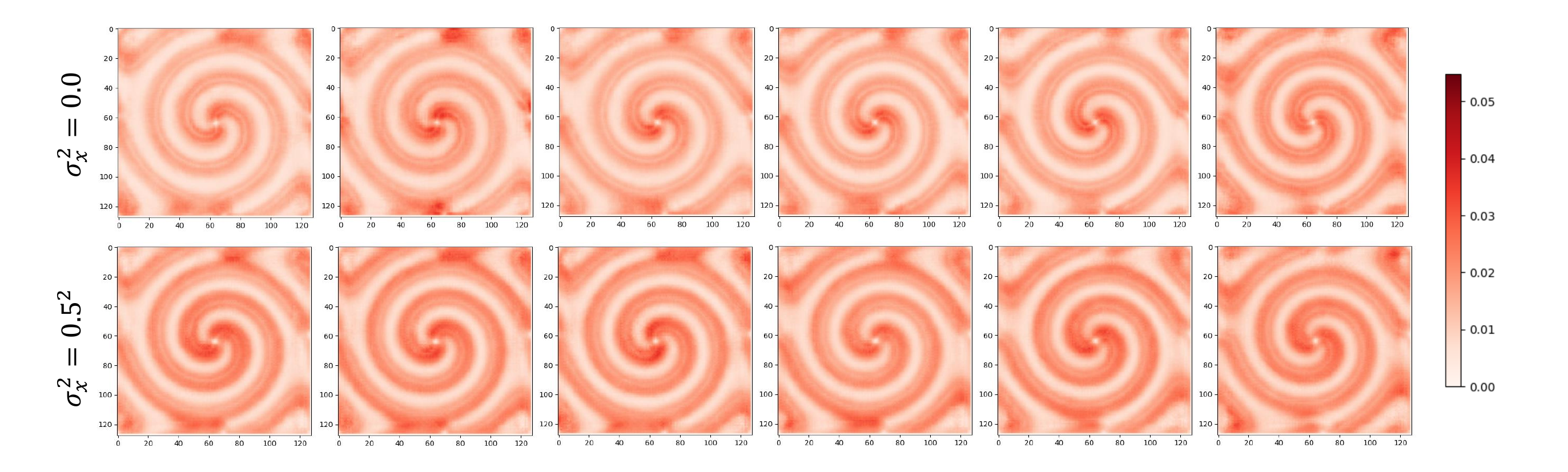}
        \caption{Evolution over time of the standard deviation of the trajectories projected in the image space for $\sigma^2_{\mathbf{x}}=0.0$ and $\sigma^2_{\mathbf{x}}=0.5^2$. }
        \label{fig:UQ_IC_det_F_sampling_T=20_0.0_PDE}
\end{figure}

In Figure \ref{fig:UQ_IC_det_F_sampling_T=20_0.0_PDE}, we show the uncertainty quantification capabilities of our framework by analyzing the uncertainties over the same trajectory with different noise levels ($\sigma^2_{\mathbf{x}}=0.0$ and $\sigma^2_{\mathbf{x}}=0.5^2$). 
Again, it is possible to notice that the model is sensitive to the noise added to the measurements, as the standard deviation of the predicted trajectory grows with more noise, i.e., the spiral is more blurred in the case of $\sigma^2_{\mathbf{x}}=0.5^2$ than in the case of $\sigma^2_{\mathbf{x}}=0.0$.

\section{Conclusion and Discussion}\label{sec:conclusion}

In this paper, we introduced a recurrent SVDKL architecture for learning ROMs for chaotic dynamical systems from high-dimensional and noisy measurements. This novel approach extends the architecture proposed in \cite{botteghi2022deep} with a recurrent network and two multi-step loss functions to improve the reliability of the long-term predictions of the latent dynamical SVDKL model. The method was tested on an actuated double pendulum and a parametric reaction-diffusion PDE. Our method, in both test cases, is capable of properly denoising the measurement, learning interpretable latent representations, and consistently predicting the evolution of the systems. Eventually, we propose a way to visualize and analyze the uncertainties quantification capabilities of the framework.

 In recent years, model and dimensionality reduction \cite{schmid2010dynamic, brunton2019data, ghattas2021learning,  peherstorfer2016data, qian2020lift, fresca2022pod,lee2020model,otto2019linearly}, uncertainty quantification  \cite{sudret2000stochastic, bui2008parametric, galbally2010non, guo2022bayesian}, and measurement denoising \cite{guo2019data, owhadi2019kernel} have been essential aspects of the research in scientific machine learning. However, novel methods have been very often tailored for only one or two of these challenges at a time. Simultaneously tackling all these challenges is very difficult. To the best of our knowledge, the proposed method in this work is the first one capable of denoising high-dimensional measurements, reducing their dimensionality into interpretable latent spaces, predicting system evolution, and quantifying modeling uncertainties simultaneously, as shown in Section \ref{sec:numericalresults}.

\section*{Acknowledgments}
AM acknowledges the Project “Reduced Order Modeling and Deep Learning for the real- time approximation of PDEs (DREAM)” (Starting Grant No. FIS00003154), funded by the Italian Science Fund (FIS) - Ministero dell'Università e della Ricerca and the project FAIR (Future Artificial Intelligence Research), funded by the NextGenerationEU program within the PNRR-PE-AI scheme (M4C2, Investment 1.3, Line on Artificial Intelligence). AM and PZ are members of the Gruppo Nazionale Calcolo Scientifico-Istituto Nazionale di Alta Matematica (GNCS-INdAM) and acknowledge the project “Dipartimento di Eccellenza” 2023-2027, funded by MUR. This work was in part carried out when MG held a position at the University of Twente (NL), for which he acknowledges financial support from \emph{Sectorplan Bèta} under the focus area \emph{Mathematics of Computational Science}.

\bibliographystyle{unsrt}  
\bibliography{references}  

\end{document}